\documentclass[conference]{IEEEtran}
\IEEEoverridecommandlockouts
\usepackage{cite}
\usepackage{amsmath,amssymb,amsfonts}
\usepackage{algorithmic}
\usepackage{graphicx}
\usepackage{textcomp}
\usepackage{verbatim}
\usepackage{xcolor}
\usepackage{float}
\usepackage{caption}
\usepackage{url}
\def\BibTeX{{\rm B\kern-.05em{\sc i\kern-.025em b}\kern-.08em
    T\kern-.1667em\lower.7ex\hbox{E}\kern-.125emX}}

\begin{document}

\title{BagDINO: Multi-View Baggage Re-Identification with DINOv3}

\author{\IEEEauthorblockN{Vita Santa Barletta}
\IEEEauthorblockA{\textit{University of Bari Aldo moro} \\
\textit{Department of Computer Science}\\
Bari, Italia \\
vita.barletta@uniba.it}
\and
\IEEEauthorblockN{Danilo Caivano}
\IEEEauthorblockA{\textit{University of Bari Aldo Moro, SER\&Practices} \\
\textit{Department of Computer Science}\\
Bari, Italia \\
danilo.caivano@uniba.it}
\and
\IEEEauthorblockN{Rebecca Margiotta}
\IEEEauthorblockA{\textit{University of Bari Aldo Moro} \\
\textit{Department of Computer Science}\\
Bari, Italia \\
r.margiotta5@studenti.uniba.it}
\and
\IEEEauthorblockN{Massimiliano Morga}
\IEEEauthorblockA{\textit{SER\&Practices} \\
\textit{Spin-off of the University of Bari Aldo Moro}\\
Bari, Italia \\
m.morga@serandp.com}
\and
\IEEEauthorblockN{Davide Pio Posa}
\IEEEauthorblockA{\textit{University of Bari Aldo Moro, SER\&Practices} \\
\textit{Department of Computer Science}\\
Bari, Italia \\
d.posa@studenti.uniba.it}
}

\maketitle

\begin{abstract}
Mishandled checked baggage remains a recurrent issue in airport operations, and current recovery workflows still largely rely on tag-based tracking, which does not directly support visual identification when tag evidence is missing or unavailable.
This paper investigates baggage re-identification as an instance-level retrieval problem in a multi-camera setting, leveraging DINOv3 foundation-model representations to match a query image against a gallery of registered baggage images.
A Torchreid-style BNNeck re-identification head is placed on top of a DINOv3 backbone, and parameter-efficient adaptation is performed via LoRA. Experiments are conducted on the MVB benchmark using a progressive study that compares a fully frozen backbone against LoRA and fine-tuning strategies.
Results indicate that parameter-efficient adaptation of foundation-model features provides an effective and stable approach for multi-view baggage re-identification under limited training data.
\end{abstract}


\section{Introduction}
Despite continuous improvements in airport operations, mishandled checked baggage remains a persistent operational and customer experience issue. Recent industry reporting estimates a global mishandling rate of 6.9 bags per 1{,}000 passengers in 2023 (down from 7.6 in 2022), with delayed bags accounting for the majority of cases (77\%) and transfer-related issues representing a large fraction of delays (46\%) \cite{SITA2024BaggageIT}. Beyond passenger frustration, mishandling also translates into non-negligible operational costs: IATA (International Air Transport Association) estimates that reuniting passengers and baggage costs roughly \$100 per bag \cite{IATA753Guide2023}.

In current airport workflows, baggage identification and tracking are predominantly tied to tag-based processes, where tracking evidence must be recorded at specific checkpoints. To standardize this process and reduce mishandling, IATA Resolution 753 requires tracking at four core points: \emph{Acceptance}, \emph{Load}, \emph{Transfer}, and \emph{Arrival} \cite{IATA753Guide2023}. While such mechanisms provide essential traceability, they do not directly address scenarios in which a bag must be recognized from images (e.g., CCTV or operational cameras), or matched across cameras when tag evidence is missing, corrupted, or unavailable at the time of inspection.

At the same time, the increasing adoption of self-service and automated check-in/bag-drop solutions creates an opportunity for \emph{visual registration} under controlled conditions. In this setting, the standard identification step (tag assignment and record creation) can be complemented by capturing a small set of multi-view images of each bag in a stable environment (consistent viewpoint constraints, lighting, and background), thus producing a clean visual ``signature'' at registration time. Later, when the bag is observed in less constrained areas of the airport, the system can perform instance-level retrieval by matching a query image against a gallery of registered multi-view images to retrieve the corresponding bag identity (and hence its associated record).

Baggage ReID is challenging because many baggage items share nearly identical global appearance due to industrial manufacturing, and the discriminative evidence is frequently confined to fine-grained instance-specific cues such as stickers, surface wear, or small damages. These cues may further be affected by viewpoint changes, occlusions, motion blur, illumination differences, and background clutter. This paper investigates baggage re-identification on the MVB benchmark using foundation-model visual representations, leveraging self-supervised transformer features for cross-view matching \cite{zhang2019mvb}.

\begin{figure*}[t]
    \centering
    \includegraphics[width=\textwidth]{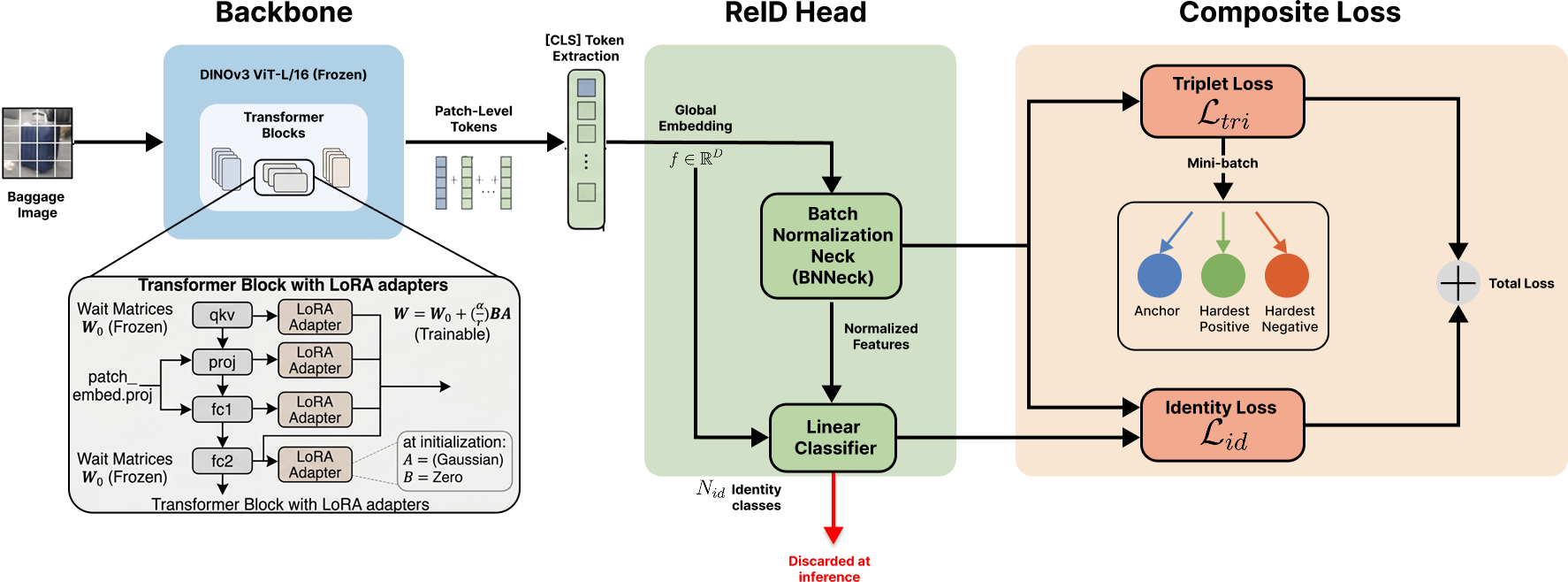}
    \caption{Overview of the proposed baggage re-identification pipeline. A DINOv3 ViT backbone provides a global image representation (from the [CLS] token) and is adapted via LoRA modules injected into selected transformer projections (e.g., \texttt{qkv}, \texttt{proj}, \texttt{fc1}, \texttt{fc2}) while keeping the pre-trained weights frozen. The resulting embedding is processed by a Torchreid-style BNNeck head and optimized with a composite objective that combines identity classification (cross-entropy with label smoothing) and batch-hard triplet loss. At inference time, the linear classifier is discarded and retrieval is performed using the normalized embeddings (e.g., cosine similarity) to rank gallery items for a given query.}
    \label{fig:pipeline}
\end{figure*}

Therefore, considering the need to improve current solutions, the paper proposes \textit{BagDINO} as a foundation-model-based approach to baggage re-identification and makes the following three main contributions:
\begin{itemize}
    \item The development of a baggage re-identification pipeline (Figure \ref{fig:pipeline}) built on top of DINOv3 ViT-L/16 representations \cite{simeoni2025dinov3} and a Torchreid-style BNNeck head \cite{luo2019bagtricksstrongbaseline}, framing the task as instance-level retrieval in a multi-camera setting.
    
    \item A LoRA-based parameter-efficient adaptation configuration \cite{hu2021loralowrankadaptationlarge} is evaluated by injecting low-rank updates into the backbone projections \texttt{qkv}, \texttt{proj}, \texttt{fc1}, and \texttt{fc2}, while keeping the original backbone weights frozen. 

    \item A progressive experimental study on MVB  \cite{zhang2019mvb} compares a fully frozen backbone, LoRA adaptation, and partial/full fine-tuning under a shared training recipe, reporting performance in terms of mAP and CMC at Rank-1/5/10; across the tested configurations, results range from 0.57 mAP and 0.60 Rank-1 in the frozen setting to 0.88 mAP and 0.87 Rank-1 with LoRA (Rank-5/10: 0.96/0.98).
    
\end{itemize}
 

The paper is structured as follows: Section \ref{related_work} reviews the main research directions in baggage re-identification and related instance-retrieval approaches. Section \ref{proposed_method} presents the proposed pipeline, including the DINOv3 backbone, the LoRA-based parameter-efficient adaptation strategy, the re-identification head, and the training objective. Section \ref{experiments} describes the experimental setup on the MVB benchmark and reports a comparative evaluation of different adaptation strategies together with an analysis of retrieval errors. 
Finally, Section \ref{cocnlusion} concludes the paper and discusses possible directions for future research.

\section{Related Work}
\label{related_work}
Baggage re-identification has recently gained momentum as a practical instance-retrieval problem in airport logistics, largely driven by the introduction of the Multi-View Baggage (MVB) benchmark, a large-scale dataset with 4{,}519 identities and 22{,}660 images acquired in real multi-camera airport environments, which highlights domain shift between acquisition points and strong variations due to viewpoint, illumination, occlusions, blur, and background clutter \cite{zhang2019mvb}.Compared to person re-identification, baggage re-identification is inherently more challenging due to the high visual similarity across different baggage identities and the significant appearance variations within the same identity \cite{Chen2020}.

In fact, early and ``classical'' pipelines on MVB typically adapt person/object ReID recipes with CNN backbones (e.g., ResNet-50) and metric learning (ID loss + triplet variants), and progressively enhance the baseline by injecting context and fine-grained cues. For example, Mazzeo \emph{et al.} \cite{mazzeo2020siamese} propose an attention/multi-granularity strategy that combines global descriptors with local feature learning (e.g., masking/erasing-style mechanisms to force attention to discriminative regions) and global-context modeling, obtaining consistent gains over strong baselines by improving robustness to viewpoint changes.
Another influential direction is to explicitly model the multi-view nature of the problem at the loss/design level: QuadNet introduces a multi-view quadruplet loss to enforce cross-view compactness for the same identity while separating hard negatives both within and across views, and further strengthens discrimination with view-aware local features (e.g., CAM-guided regions), robustness-oriented augmentations (e.g., blur), and auxiliary supervision such as material prediction, reporting substantial improvements over conventional ReID baselines on MVB \cite{yang2022quadnet}.

More recent works focus on improving cross-domain and cross-camera robustness and the final ranking stage; for instance, methods that combine domain-robust feature learning (e.g., normalization and attention modules), hard-sample mining, and camera-aware re-ranking (as in CDR-CARNet) show that carefully addressing domain shift and exploiting camera structure can markedly boost mAP and Rank-1 on MVB \cite{liu2025cdr}.

The Region-to-Global Vision Transformer (RGViT), the best reported SOTA in our survey, aggregates local evidence into regional tokens and explicitly models global spatial relations between regions, optimizing the embedding with a quadruplet objective designed for near-duplicate instances; in MVB it reports 88.7\% Rank-1 and 89.6\% mAP, suggesting that structured local-to-global modeling is particularly effective when many identities share highly similar global appearance \cite{xing2023region}.

Although recent methods have significantly improved performance on the MVB benchmark, most existing solutions still rely on adaptations of person re-identification pipelines originally designed for human appearance modeling. However, baggage objects present different visual characteristics compared to human subjects, including weaker structural priors, highly similar object categories, and frequent occlusions or deformations during handling. Consequently, approaches developed for person ReID may not fully exploit the discriminative cues specific to baggage identification.

This observation highlights the need for more specialized feature learning strategies capable of capturing both global structural patterns and subtle local differences between visually similar baggage items.

\section{Proposed Method}
\label{proposed_method}
The proposed approach, \textit{BagDINO}, aims to improve robustness to viewpoint variations, background clutter, and visual similarity across baggage identities, ultimately leading to more reliable instance matching in real-world airport scenarios. 
Consequently, the proposed pipeline for baggage ReID is organized around three main components:
\begin{itemize}
    \item \emph{Backbone}: DINOv3 foundation model with parameter-efficient fine-tuning via Low-Rank Adaptation;
    \item \emph{ReID Head}: takes the backbone patch token and produces the final embeddings;
    \item \emph{Composite Loss}: a combination of a Triplet loss and a Cross-Entropy loss used during training.
\end{itemize}

\subsection{Backbone and LoRA Fine-Tuning}
As visual backbone we adopt DINOv3 \cite{simeoni2025dinov3}, a Vision Transformer (ViT) pre-trained with self-supervised objectives on a large and curated image corpus. DINOv3 produces rich, semantically meaningful patch-level representations that have shown strong transferability across visual tasks without task-specific architectural changes.

Given the relatively small scale of the MVB training split and the high computational cost of full fine-tuning a large ViT, we adopt Low-Rank Adaptation (LoRA) \cite{hu2021loralowrankadaptationlarge} as a parameter-efficient fine-tuning strategy. LoRA injects trainable low-rank decomposition matrices into selected linear projections of each Transformer
block, while keeping the original pre-trained weights frozen.

Formally, for a pre-trained weight matrix
\(\mathbf{W}_0 \in \mathbb{R}^{d \times k}\), LoRA parameterizes the update as:
\[
\mathbf{W} = \mathbf{W}_0 + \Delta\mathbf{W} = \mathbf{W}_0 +
\frac{\alpha}{r}\mathbf{B}\mathbf{A},
\]
where \(\mathbf{W}_0\) is kept frozen during training,
\(\mathbf{A} \in \mathbb{R}^{r \times k}\) and
\(\mathbf{B} \in \mathbb{R}^{d \times r}\) are the trainable low-rank matrices,
\(r \ll \min(d,k)\) is the rank hyperparameter controlling the number of
trainable parameters, and \(\alpha\) is a scalar that governs the magnitude
of the adaptation \cite{hu2021loralowrankadaptationlarge}.
At initialization, \(\mathbf{B} = \mathbf{0}\) and \(\mathbf{A}\) is randomly
drawn from a Gaussian distribution, ensuring \(\Delta\mathbf{W} = \mathbf{0}\)
at the start of fine-tuning \cite{hu2021loralowrankadaptationlarge}.

In this implementation, LoRA adapters are applied to four linear projections within each Transformer block: the combined Query-Key-Value projection (\texttt{qkv}), the attention output projection (\texttt{proj}), and the two fully-connected layers of the feed-forward network (\texttt{fc1}, \texttt{fc2}). This choice ensures that both the attention mechanism and the token-wise non-linear transformations are adapted to the baggage domain, while the pre-trained weights of the backbone remain frozen.

\subsection{Re-Identification Head}

\begin{figure}[H]
    \centering
    \includegraphics[width=\columnwidth]{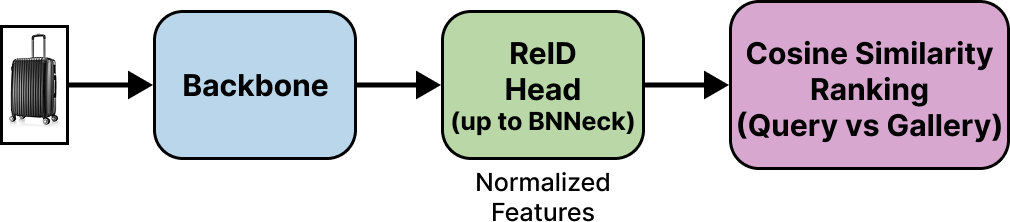}
    \caption{Inference-stage retrieval. Query and gallery images are embedded by the backbone+BNNeck, and gallery instances are ranked by similarity in the embedding space to produce the final match list.}
    \label{fig:pipeline_inference}
\end{figure}

The backbone produces a sequence of patch tokens, so the \emph{[CLS]} token representation is extracted as the global image descriptor, as it aggregates information from all spatial positions through the self-attention mechanism.
This embedding \(\mathbf{f} \in \mathbb{R}^{D}\) is then passed through the
re-identification head, which consists of:

\begin{enumerate}
    \item \emph{Batch Normalization Neck} (BNNeck) \cite{luo2019bagtricksstrongbaseline}: a batch normalization layer applied after the backbone output and before the classification layer. Following common ReID practice, this design separates the feature representation used for metric learning from the one used for identity classification, which is known to support stable optimization and effective retrieval performance \cite{luo2019bagtricksstrongbaseline}. 

    \item \emph{Linear Classifier}: a fully-connected layer with \(N_\text{id}\) outputs (one per training identity), used only during training to compute the identity loss; at inference time this layer is discarded and retrieval is
    performed directly via cosine similarity (Figure \ref{fig:pipeline_inference}).
\end{enumerate}

At inference time, given a query image and a gallery of images, the gallery items are ranked by decreasing cosine similarity:
\[
s(\mathbf{x}_q, \mathbf{x}_i) = \frac{\mathbf{f}_q \cdot \mathbf{f}_i}{\|\mathbf{f}_q\|\,\|\mathbf{f}_i\|}
\]

\subsection{Loss Function}
Training is supervised by a combination of an identity classification loss and a metric learning loss, following the standard approach in ReID pipelines \cite{luo2019bagtricksstrongbaseline}. 

\paragraph{Identity Loss}

Identity classification loss is a standard cross-entropy loss with label smoothing \cite{szegedy2016rethinking} applied to post-BN features \(\hat{\mathbf{f}}\):
\[
\mathcal{L}_{\text{ID}} =
- \sum_{c=1}^{N_{\text{id}}}
  \tilde{y}_c \log \hat{p}_c,
\]

where \(\hat{p}_c\) is the predicted probability for class \(c\) and \(\tilde{y}_c = (1-\varepsilon)\,y_c + \varepsilon / N_{\text{id}}\) is the smoothed label, with \(\varepsilon\) the smoothing factor.
Label smoothing regularizes the classifier by preventing overconfident predictions on training identities.

\paragraph{Triplet Loss}
The metric learning component is a batch-hard triplet loss \cite{hermans2017defensetripletlossperson}, which, for each anchor sample in a mini-batch, selects the hardest positive (same identity, maximum distance) and the hardest negative (different identity, minimum distance):
\[
\mathcal{L}_{\text{Tri}} =
\sum_{a \in \mathcal{B}}
\left[
  \max_{p:\,y_p=y_a} d(\mathbf{f}_a, \mathbf{f}_p)
  -
  \min_{n:\,y_n \neq y_a} d(\mathbf{f}_a, \mathbf{f}_n)
  + m
\right]_{+},
\]
where \(d(\cdot,\cdot)\) denotes Euclidean distance, \(m\) is the margin, and \([\,\cdot\,]_{+} = \max(0,\cdot)\). \\

\paragraph{Total Loss}

The overall training objective is the weighted sum of the two terms:
\[
\mathcal{L} = \mathcal{L}_{\text{ID}} + \lambda\,\mathcal{L}_{\text{Tri}},
\]
where \(\lambda\) controls the relative contribution of the metric learning term with respect to the identity classification term.

\section{Experiments}
\label{experiments}

\subsection{MVB Dataset}

All experiments are conducted on the Multi-View Baggage (MVB) benchmark, a large-scale dataset designed for baggage re-identification in real airport environments \cite{zhang2019mvb}. MVB contains 4{,}519 baggage identities and 22{,}660 annotated baggage images, and provides additional annotations such as surface material labels (e.g. soft, hard) and segmentation masks in the original release. 
Importantly, MVB is not limited to standard suitcases (e.g. rigid trolleys), but also includes a broader range of baggage items such as soft bags, duffels and transported parcels, better reflecting the diversity of objects that may be checked in and subsequently matched across cameras.
Images are collected with a specially designed multi-view camera system and span two acquisition points with substantially different imaging condition, commonly described as BHS/conveyor-belt views and checkpoint views. This introduces a realistic domain gap across acquisition points (background, lighting, viewpoint/pose), and the dataset also exhibits strong intra-class variation due to occlusions, motion blur and background clutter. 
As a result, MVB does not represent a fully controlled capture setup, even though each acquisition point can be interpreted as an operational checkpoint that is compatible with the visual registration setting discussed in industry workflows. All experiments follow the official train/query/gallery splits and the evaluation protocol defined in the original paper. Performance is reported using standard retrieval metrics, namely mean Average Precision (mAP) and Cumulative Matching Characteristics (CMC) at Rank-1, Rank-5, and Rank-10.

\begin{figure}[t]
    \centering
    \includegraphics[width=\columnwidth]{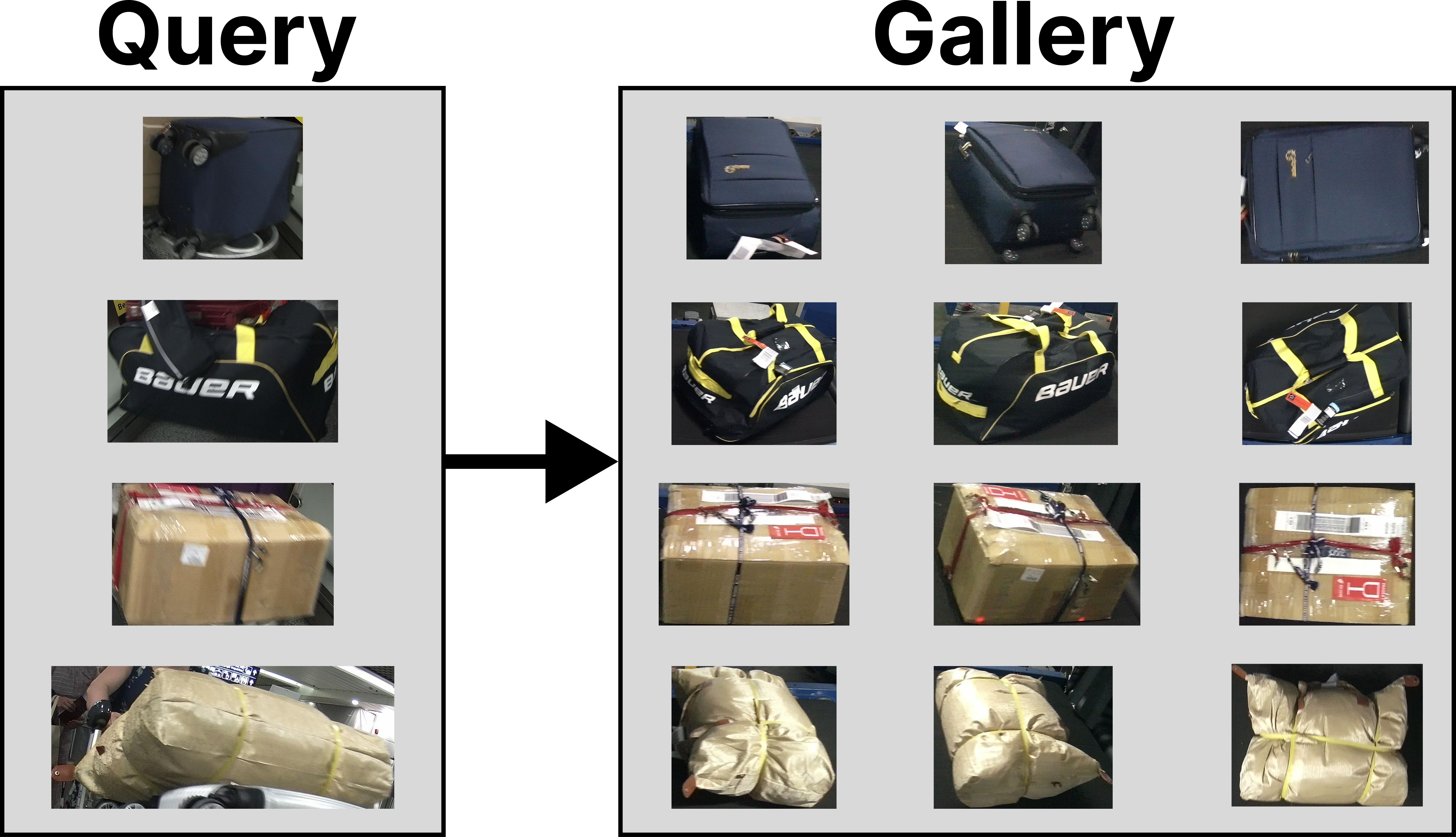}
    \caption{Example MVB query-gallery pairs illustrating the heterogeneity of baggage items and the variability induced by different acquisition conditions.}
    \label{fig:mvb_examples}
\end{figure}

\subsection{Experimental Setup}
All experiments follow a Torchreid-style training recipe with a BNNeck head and a composite objective that combines identity classification (cross-entropy) and batch-hard triplet loss (margin $m=0.3$) with equal weights. We resize all images to $256 \times 256$ and apply standard ImageNet normalization; data augmentation includes random horizontal flipping ($p=0.5$) and random cropping with padding (10 pixels). Training uses PK sampling with batch size 32 and $K=4$ instances per identity (i.e., $P=8$ identities per mini-batch) for 80 epochs, with the random seed fixed to 42. Optimization is performed with AdamW (base learning rate $3.25 \times 10^{-4}$, weight decay $5 \times 10^{-4}$) and a cosine learning-rate schedule; we use separate optimizer parameter groups with a 10$\times$ smaller learning rate for trainable LoRA parameters in the backbone than for the BNNeck head. We apply gradient clipping with max norm 5.0 and optionally enable early stopping (patience 15, minimum $\Delta$ of 0.001). For LoRA-based adaptation, low-rank updates (rank 16, $\alpha=32$) are injected into the backbone projections \texttt{qkv}, \texttt{proj}, \texttt{fc1}, and \texttt{fc2}.

\subsection{Progressive Study}
A progressive experimental analysis is conducted to quantify the impact of parameter-efficient adaptation and to compare it with partial and full fine-tuning strategies, while keeping the same ReID head and loss formulation across all runs.

\paragraph{Frozen backbone}
To evaluate the out-of-the-box transferability of DINOv3 features to the baggage ReID task, the first experiment uses a fully frozen DINOv3 ViT-L/16 backbone and trains only the ReID head. This baseline achieves 0.57 mAP, with Rank-1/5/10 equal to 0.60/0.80/0.87, outperforming the original baseline reported in the MVB paper \cite{zhang2019mvb}. These results confirm that foundation-model representations provide a strong starting point, although domain-specific adaptation remains necessary to better cope with the cross-view variability of MVB.

\paragraph{LoRA adaptation}
Low-Rank Adaptation (LoRA) is then introduced to specialize the backbone to the operational baggage domain. LoRA modules are applied to \texttt{qkv}, \texttt{proj}, \texttt{fc1}, and \texttt{fc2} layers (rank 16, alpha 32). This configuration yields a substantial performance improvement, reaching 0.88 mAP with Rank-1/5/10 of 0.87/0.96/0.98. The gain indicates that parameter-efficient updates are sufficient to adapt high-capacity ViT representations to fine-grained baggage cues, while mitigating the overfitting risk typically associated with full fine-tuning on limited training data.

\paragraph{Partial fine-tuning}
Although full fine-tuning is not the primary strategy due to the limited size of the training split, a partial fine-tuning setting is evaluated for completeness. In this configuration, 75\% of the backbone remains frozen, while the remaining layers are updated using a conservative learning rate (on the order of \(10^{-6}\)) and stronger weight decay. The model achieves 0.83 mAP with Rank-1/5/10 of 0.83/0.95/0.97. While competitive, this setup underperforms LoRA, suggesting that even partial fine-tuning may require more data to effectively adapt the backbone without degrading its generalization properties.

\paragraph{Full fine-tuning}
Finally, all backbone parameters are unfrozen and optimized under the same conservative regime (low learning rate and aggressive weight decay). This configuration achieves 0.85 mAP, with Rank-1/5/10 equal to 0.86/0.95/0.96. Overall, the fine-tuning experiments confirm the initial hypothesis: given the restricted training size and the high visual similarity among many MVB instances, LoRA provides a more stable and effective adaptation mechanism than updating the entire backbone.

\subsection{Error Analysis}
\begin{figure}[t]
    \centering
    \includegraphics[width=\columnwidth]{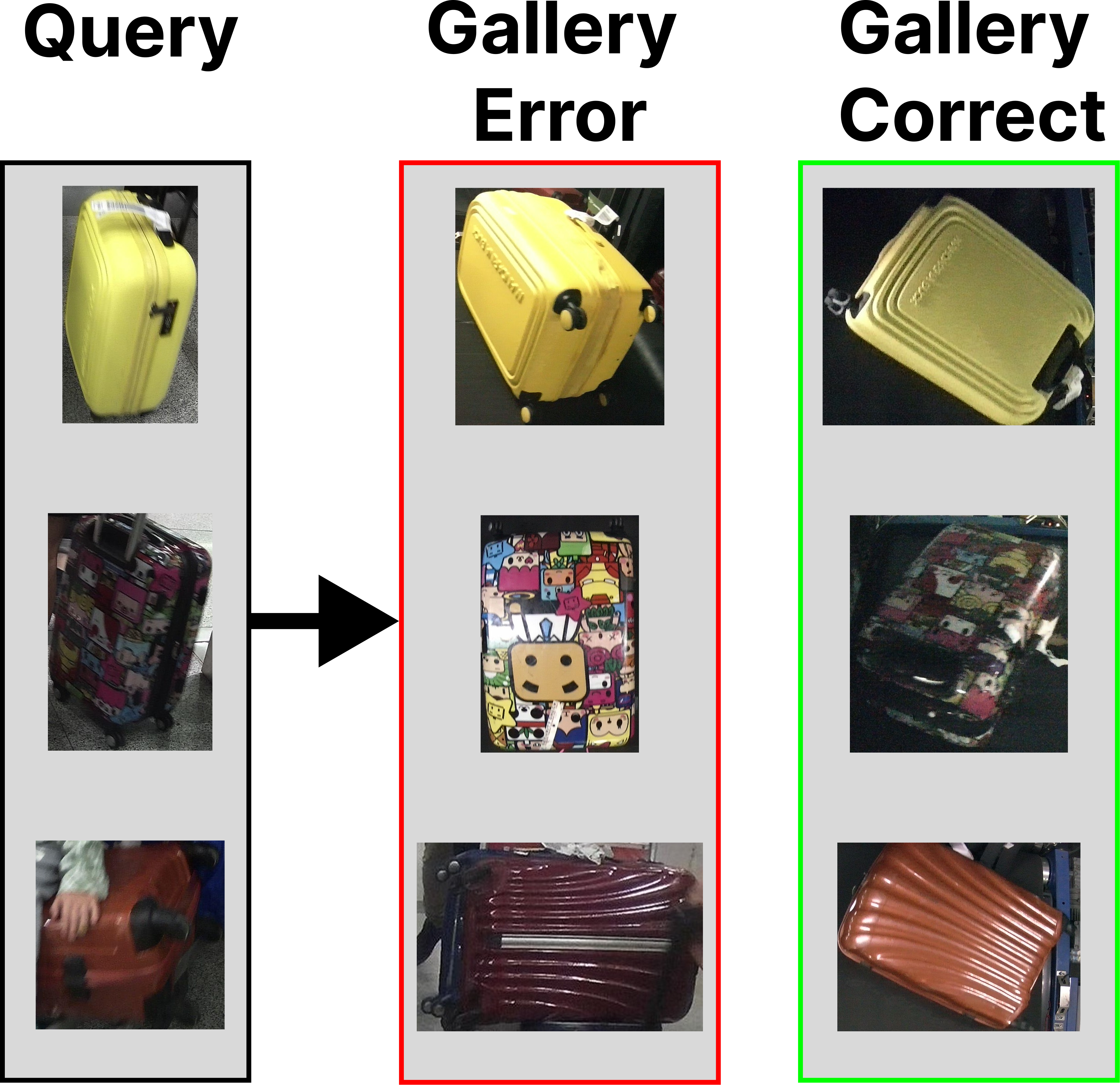}
    \caption{High-similarity failure cases on MVB. Three examples are reported, each shown as (i) the query, (ii) the corresponding true match in the gallery (positive), and (iii) the top-ranked incorrect retrieval returned with high confidence (similarity $\gtrsim 78\%$). These errors typically occur when different baggage identities exhibit near-duplicate visual cues (e.g., similar shape, material, and color patterns), yielding very close neighbors in the embedding space.}
    \label{fig:hard_errors}
\end{figure}

Beyond standard retrieval metrics, the purity of the top-$k$ list is inspected, i.e., the fraction of the first $k$ retrieved items that match the query identity. Under the adopted test protocol, each identity is represented by only a small number of gallery images (typically 1-3), which makes purity@5 inherently conservative: even when all available true matches are retrieved at the top, the remaining positions in the top-5 list must be filled with other identities and are therefore counted as impurities. 
In an operational setting, visual registration would typically store a larger multi-view template per item (e.g., 6+ images), increasing the number of available true matches and yielding a purer shortlist of candidates at fixed $k$. Consistently with the high Rank-1 performance of the best configuration (0.87), failures are relatively infrequent; however, when they occur, they often correspond to visually near-duplicate baggage items in the gallery (similar shape, material, and color patterns), leading to incorrect matches with high similarity scores. Figure \ref{fig:hard_errors} reports representative high-similarity failure cases, where the top-ranked gallery image is incorrect despite being very close to the query in the embedding space.

\subsection{Results Summary}

Table~\ref{tab:results_summary_ablation} summarizes the adaptation strategies evaluated on MVB. A frozen BagDINO backbone already provides a strong baseline (0.57 mAP, 0.60 Rank-1), while LoRA-based adaptation yields the best overall trade-off among the tested settings (0.88 mAP, 0.87 Rank-1). Partial and full fine-tuning remain competitive but do not match the LoRA configuration under the same training recipe.

Table~\ref{tab:results_summary_sota} compares BagDINO + LoRA with representative methods from the literature on the same benchmark. The proposed approach is competitive with prior work and achieves results close to the strongest published methods, while remaining based on a parameter-efficient adaptation scheme.

More generally, the results should also be interpreted in light of the different design choices adopted by the compared approaches. In particular, while some methods, such as RGViT \cite{xing2023region}, introduce additional architectural modules to improve feature modeling, BagDINO relies on a parameter-efficient LoRA configuration, where only a limited set of low-rank updates is optimized while the DINOv3 backbone remains frozen. This suggests that the observed differences across methods are not only related to benchmark scores but also reflect different trade-offs between architectural specialization and lightweight adaptation.

\begin{table}[H]
\centering
\begin{tabular}{lcc}
\hline
\textbf{Method} & \textbf{mAP} & \textbf{Rank-1} \\
\hline
BagDINO (Frozen)   & 0.57 & 0.60 \\
\textbf{BagDINO + LoRA}     & \textbf{0.88} & \textbf{0.87} \\
BagDINO FT (25\%)  & 0.83 & 0.83 \\
BagDINO FT (100\%) & 0.83 & 0.83 \\
\hline
\end{tabular}

\caption{Results on the MVB benchmark in terms of mAP and Rank-1 summarizing the evaluated DINOv3-based configurations under different adaptation strategies.}
\label{tab:results_summary_ablation}
\end{table}

\begin{table}[H]
\centering
\begin{tabular}{lcc}
\hline
\textbf{Architecture} & \textbf{mAP} & \textbf{Rank-1} \\
\hline
Huangbin et al. \cite{wu2021attention} & 0.83 & 0.85 \\
Yang et al.     \cite{yang2022quadnet} & 0.86 & 0.88 \\
Liu et al.      \cite{liu2025cdr}      & 0.87 & 0.86 \\
Zhao et al.     \cite{zhao2021mvad}    & \textbf{0.91} & 0.87 \\
Mazzeo et al.   \cite{mazzeo2020siamese} & 0.63 & 0.69 \\
Zhiwei et al.   \cite{xing2023region}  & 0.90 & \textbf{0.89} \\
Zhang et al.    \cite{zhang2019mvb}    & 0.50 & 0.50 \\
BagDINO + LoRA     & 0.88 & 0.87 \\
\hline
\end{tabular}

\caption{Results on the MVB benchmark in terms of mAP and Rank-1 comparing BagDINO best approach with the representative methods from literature.}
\label{tab:results_summary_sota}
\end{table}

\paragraph{Cross-Domain Evaluation}
To assess whether the proposed approach can transfer beyond the baggage domain, we additionally evaluate it on Market-1501 \cite{zheng2015market1501}. Market-1501 is a widely used person re-identification benchmark composed of pedestrian images acquired from multiple cameras, and it is commonly adopted to evaluate cross-camera retrieval performance. This experiment is not intended as an in-domain comparison with MVB, but rather as an out-of-domain validation to assess whether the proposed adaptation strategy remains effective in a different visual context.

Table~\ref{tab:market1501_transfer} reports the comparison with baggage ReID methods that were also benchmarked on Market-1501, where BagDINO + LoRA achieves the highest mAP and Rank-1 among the considered approaches.

\begin{table}[H]
\centering
\begin{tabular}{lcc}
\hline
\textbf{Method} & \textbf{mAP} & \textbf{Rank-1} \\
\hline
BagDINO + LoRA & \textbf{93.63} & \textbf{96.32} \\
Yang et al. \cite{yang2022quadnet} & 88.60 & 95.40 \\
Mazzeo et al. \cite{mazzeo2020siamese} & 80.92 & 85.79 \\
\hline
\end{tabular}
\caption{Cross-domain evaluation on Market-1501 to assess the transferability of the proposed approach beyond the baggage domain.}
\label{tab:market1501_transfer}
\end{table}

\section{Conclusions}
\label{cocnlusion}

This work investigated baggage re-identification as an instance-level retrieval problem in multi-camera airport environments, leveraging foundation-model representations to match query images against a gallery of registered baggage views. A DINOv3 ViT backbone combined with a Torchreid-style BNNeck head was evaluated under different adaptation strategies, including a fully frozen backbone, parameter-efficient LoRA adaptation, and partial/full fine-tuning. On the MVB benchmark, LoRA-based adaptation provided the strongest overall results among the tested configurations, reaching 0.88 mAP and 0.87 Rank-1, while maintaining a stable training regime under limited training data.

The qualitative inspection of retrieval lists highlighted two practical aspects of the benchmark setting: the limited number of gallery images per identity in the test split makes purity at fixed top-$k$ inherently conservative, and the remaining failures are typically associated with visually near-duplicate baggage items that produce high-similarity confusions. These observations suggest that future evaluations aligned with operational deployment could benefit from larger multi-view templates per item at registration time, as well as from additional mechanisms for handling near-duplicate instances (e.g., stronger fine-grained cues, re-ranking, or auxiliary supervision) without compromising retrieval efficiency.

\section*{Acknowledgment}
This work was partially supported by the following project: ``ThermoNET''  - Codice CUP B25H25002090007. TITOLO II CAPO 2 DEL REGOLAMENTO GENERALE “Aiuti ai programmi integrati promossi da MEDIE IMPRESE ai sensi dell’articolo 26 del Regolamento”; Patto territoriale "Sistema universitario pugliese" – CUP F61B23000370006.

\bibliographystyle{ieeetr}
\bibliography{bibliography}

@report{SITA2024BaggageIT,
  author      = {{SITA}},
  title       = {SITA Baggage IT Insights 2024},
  institution = {SITA},
  year        = {2024},
  url         = {https://www.sita.aero/resources/surveys-reports/sita-baggage-it-insights-2024/},
}

@manual{IATA753Guide2023,
  author       = {{International Air Transport Association (IATA)}},
  title        = {Baggage Tracking -- IATA Resolution 753 Implementation Guide},
  year         = {2023},
  edition      = {Issue 4.0},
  month        = nov,
  organization = {International Air Transport Association},
  url          = {https://www.iata.org/contentassets/5c4aa8b8b3b1432697d2bf3301450684/reso753-implementation-guide---2023_issue-4.0.pdf},
}

@inproceedings{zhang2019mvb,
  title={MVB: A large-scale dataset for baggage re-identification and merged Siamese networks},
  author={Zhang, Zhulin and Li, Dong and Wu, Jinhua and Sun, Yunda and Zhang, Li},
  booktitle={Chinese Conference on Pattern Recognition and Computer Vision (PRCV)},
  pages={84--96},
  year={2019},
  organization={Springer}
}

@article{mazzeo2020siamese,
  title={A siamese neural network for non-invasive baggage re-identification},
  author={Mazzeo, Pier Luigi and Libetta, Christian and Spagnolo, Paolo and Distante, Cosimo},
  journal={Journal of Imaging},
  volume={6},
  number={11},
  pages={126},
  year={2020},
  publisher={MDPI}
}

@article{yang2022quadnet,
  title={QuadNet: Quadruplet loss for multi-view learning in baggage re-identification},
  author={Yang, Hao and Chu, Xiuxiu and Zhang, Li and Sun, Yunda and Li, Dong and Maybank, Stephen J},
  journal={Pattern Recognition},
  volume={126},
  pages={108546},
  year={2022},
  publisher={Elsevier}
}

@article{liu2025cdr,
  title={CDR-CARNet: Baggage re-identification based on cross-domain robust features and camera-aware re-ranking},
  author={Liu, Yinghong and Zhang, Hongying and Yang, Xi and Zhao, Sijia and Zhang, Jinhong},
  journal={Computers \& Graphics},
  pages={104377},
  year={2025},
  publisher={Elsevier}
}

@inproceedings{xing2023region,
  title={Region to Global Vision Transformer for Baggage Re-Identification},
  author={Xing, Zhiwei and Zhu, Shujie and Zhang, Tao and Luo, Qian},
  booktitle={2023 42nd Chinese Control Conference (CCC)},
  pages={7433--7439},
  year={2023},
  organization={IEEE}
}

@misc{simeoni2025dinov3,
  title={{DINOv3}},
  author={Sim{\'e}oni, Oriane and Vo, Huy V. and Seitzer, Maximilian and Baldassarre, Federico and Oquab, Maxime and Jose, Cijo and Khalidov, Vasil and Szafraniec, Marc and Yi, Seungeun and Ramamonjisoa, Micha{\"e}l and Massa, Francisco and Haziza, Daniel and Wehrstedt, Luca and Wang, Jianyuan and Darcet, Timoth{\'e}e and Moutakanni, Th{\'e}o and Sentana, Leonel and Roberts, Claire and Vedaldi, Andrea and Tolan, Jamie and Brandt, John and Couprie, Camille and Mairal, Julien and J{\'e}gou, Herv{\'e} and Labatut, Patrick and Bojanowski, Piotr},
  year={2025},
  eprint={2508.10104},
  archivePrefix={arXiv},
  primaryClass={cs.CV},
  url={https://arxiv.org/abs/2508.10104},
}

@misc{hu2021loralowrankadaptationlarge,
      title={LoRA: Low-Rank Adaptation of Large Language Models}, 
      author={Edward J. Hu and Yelong Shen and Phillip Wallis and Zeyuan Allen-Zhu and Yuanzhi Li and Shean Wang and Lu Wang and Weizhu Chen},
      year={2021},
      eprint={2106.09685},
      archivePrefix={arXiv},
      primaryClass={cs.CL},
      url={https://arxiv.org/abs/2106.09685}, 
}

@misc{luo2019bagtricksstrongbaseline,
      title={Bag of Tricks and A Strong Baseline for Deep Person Re-identification}, 
      author={Hao Luo and Youzhi Gu and Xingyu Liao and Shenqi Lai and Wei Jiang},
      year={2019},
      eprint={1903.07071},
      archivePrefix={arXiv},
      primaryClass={cs.CV},
      url={https://arxiv.org/abs/1903.07071}, 
}

@misc{hermans2017defensetripletlossperson,
      title={In Defense of the Triplet Loss for Person Re-Identification}, 
      author={Alexander Hermans and Lucas Beyer and Bastian Leibe},
      year={2017},
      eprint={1703.07737},
      archivePrefix={arXiv},
      primaryClass={cs.CV},
      url={https://arxiv.org/abs/1703.07737}, 
}

@inproceedings{szegedy2016rethinking,
  title={Rethinking the inception architecture for computer vision},
  author={Szegedy, Christian and Vanhoucke, Vincent and Ioffe, Sergey and Shlens, Jon and Wojna, Zbigniew},
  booktitle={Proceedings of the IEEE conference on computer vision and pattern recognition},
  pages={2818--2826},
  year={2016}
}

@inproceedings{wu2021attention,
  title={Attention and Multi-granied Feature Learning for Baggage Re-identification},
  author={Wu, Huangbin and Luo, Zhiming and Cao, Donglin and Lin, Dazhen and Su, Songzhi and Li, Shaozi},
  booktitle={CCF Conference on Computer Supported Cooperative Work and Social Computing},
  pages={460--472},
  year={2021},
  organization={Springer}
}

@inproceedings{zhao2021mvad,
  title={MVAD-Net: Learning View-Aware and Domain-Invariant Representation for Baggage Re-identification},
  author={Zhao, Qing and Ma, Huimin and Lu, Ruiqi and Chen, Yanxian and Li, Dong},
  booktitle={Chinese Conference on Pattern Recognition and Computer Vision (PRCV)},
  pages={142--153},
  year={2021},
  organization={Springer}
}

@INPROCEEDINGS{Chen2020,
  author={Chen, Ruibo and Zhang, Hui and Li, Chen and Wang, Yaonan},
  booktitle={2020 Chinese Automation Congress (CAC)}, 
  title={LSDNN: Local-Salient Deep Neural Network for Baggage Re-Identification with Material Discerning}, 
  year={2020},
  volume={},
  number={},
  pages={6344-6349},
  doi={10.1109/CAC51589.2020.9327737}}

@InProceedings{zheng2015market1501,
  author    = {Zheng, Liang and Shen, Liyue and Tian, Lu and Wang, Shengjin and Wang, Jingdong and Tian, Qi},
  title     = {Scalable Person Re-Identification: A Benchmark},
  booktitle = {Proceedings of the IEEE International Conference on Computer Vision (ICCV)},
  year      = {2015},
  month     = {December}
}

\end{document}